\documentclass[conference]{IEEEtran}

\newcommand{\argmax}{\operatornamewithlimits{argmax}}

\usepackage{times}
\usepackage{graphicx}
\usepackage{amsmath}
\usepackage{algorithm}
\usepackage{algorithmic}
\usepackage{url}
\usepackage{amsfonts}

\begin{document}

\title{\ \\ \LARGE\bf Reinforcement Learning for the Soccer Dribbling Task \thanks{Arthur Carvalho is with the David Cheriton School of Computer Science at the University of Waterloo, Waterloo, Ontario, Canada (email: {\tt a3carval@cs.uwaterloo.ca}).}   
                \thanks{Renato Oliveira is with the Center of Informatics at the Federal University of Pernambuco, Recife, Pernambuco, Brazil (email: {\tt rmo@cin.ufpe.br}).} }

\author{Arthur Carvalho and Renato Oliveira}

\maketitle

\begin{abstract}
We propose a reinforcement learning solution to the \emph{soccer dribbling task}, a scenario in which a soccer agent has to go from the beginning to the end of a region keeping possession of the ball, as an adversary attempts to gain possession. While the adversary uses a stationary policy, the dribbler learns the best action to take at each decision point. After defining meaningful variables to represent the state space, and high-level macro-actions to incorporate domain knowledge, we describe our application of the reinforcement learning algorithm \emph{Sarsa} with CMAC for function approximation. Our experiments show that, after the training period, the dribbler is able to accomplish its task against a strong adversary around $58\%$ of the time. 
\end{abstract}

\section{Introduction}

Soccer dribbling consists of the ability of a soccer agent to go from the beginning to the end of a region keeping possession of the ball, while an adversary attempts to gain possession. In this work, we focus on the dribbler's learning process, \textit{i.e.}, the learning of an effective policy that determines\break a good action for the dribbler to take at each decision point.

We study the soccer dribbling task using the RoboCup soccer simulator~\cite{Noda:soccerserver}. Specific details of this simulator increase the complexity of the learning process. For example, besides the adversarial and real-time environment, agents' perceptions and actions are noisy and asynchronous.

We model the soccer dribbling task as a \emph{reinforcement learning} problem. Our solution to this problem combines the Sarsa algorithm with CMAC for function approximation.  Despite the fact that the resulting learning algorithm is not guaranteed to converge to the optimal policy in all cases, many lines of evidence suggest that it converges to near-optimal policies (for example, see~\cite{Gordon:Reinforcement_Learning, Sutton:generalizationin, Tsitsiklis:Function_Approximation, Perkins:Convergence}).

Besides this introductory section, the rest of this paper is organized as follows. In the next section, we describe the soccer dribbling task. In Section 3, we show how to map this task onto an episodic reinforcement learning framework. In Section 4 and 5, we present, respectively, the reinforcement learning algorithm and its results against a strong adversary. In Section 6, we review the literature related to our work. In Section 7, we conclude and present future research directions.

\section{Soccer Dribbling}

Soccer dribbling is a crucial skill for an agent to become a successful soccer player. It consists of the ability of a soccer agent, henceforth called the \emph{dribbler}, to go from the beginning to the end of a region keeping possession of the ball, while an adversary attempts to gain possession. We can see  soccer dribbling as a subproblem of the complete soccer domain. The main simplification is that the players involved are only focused on specific goals, without worrying about team strategies or unrelated individual skills (\textit{e.g.}, passing and shooting). Nevertheless, a successful policy learned by the dribbler can be used in the complete soccer domain whenever a soccer agent faces a dribbling situation.

Since our focus is on the dribbler's learning process, an omniscient coach agent is used to manage the play. At the beginning of each trial (\emph{episode}), the coach resets the location of the ball and of the players within a \emph{training field}. The dribbler is placed in the center-left region together with the ball. The adversary is placed in a random position with the constraint that it does not start with possession of the ball. An example of a starting configuration is shown in Figure 1.

\begin{figure}[b]
\centering
\caption{Example of a starting configuration.}
\includegraphics[width=0.75\columnwidth]{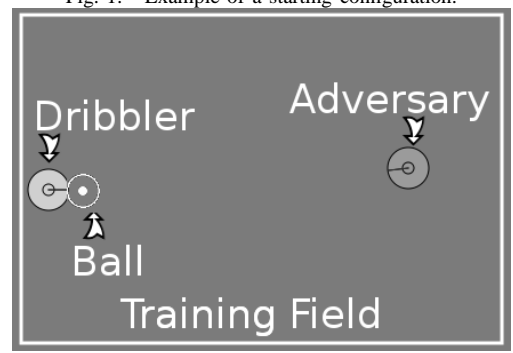}
\end{figure}

Whenever the adversary gains possession for a set period of time or when the ball goes out of the training field by crossing either the left line or the top line or the bottom line, the coach declares the adversary as the winner of the episode. If the ball goes out of the training field by crossing the right line, then the winner is the first player to intercept the ball. After declaring the winner of an episode, the coach resets the location of the players and of the ball within the training field and starts a new episode. Thus, the dribbler's goal is to reach the right line that delimits the training field with the ball. We call this task the \emph{soccer dribbling task}.

We argue that the soccer dribbling task is an excellent benchmark for comparing different machine learning techniques since it involves a complex problem, and it has a well-defined objective, which is to maximize the number of episodes won by the dribbler. We study the soccer dribbling task using the RoboCup soccer simulator~\cite{Noda:soccerserver}.

The RoboCup soccer simulator operates in discrete time steps, each  representing 100 milliseconds of simulated time. Specific details of this simulator increase the complexity of the learning process. For example, random noise is injected into all perceptions and actions. Further, agents must sense and act asynchronously. Each soccer agent receives visual information about other objects every 150 milliseconds, \textit{e.g.}, its distance from other players in its current field of view. Each agent has also a body sensor, which detects its current ``physical status" every 100 milliseconds, \textit{e.g.}, that agent's stamina and speed. Agents may execute a parameterized primitive action every 100 milliseconds, \textit{e.g.}, \emph{turn}(angle), \emph{dash}(power), and \emph{kick}(power, angle). Full details of the RoboCup soccer simulator are presented by Chen \textit{et al.}~\cite{Manual_RoboCup}.

Since possession is not well-defined in the RoboCup soccer simulator, we consider that an agent has possession of the ball whenever the ball is close enough to be kicked, \textit{i.e.}, it is in a distance less than $1.085$ meters from the agent.

\section{The Soccer Dribbling Task as a Reinforcement Learning Problem}

In the soccer dribbling task, an episode begins when the dribbler may take the first action. When an episode ends (\textit{e.g.}, when the adversary gains possession for a set period of time), the coach starts a new one, thereby giving rise to a series of episodes. Thus, the interaction between the dribbler and the environment naturally breaks down into a sequence of distinct episodes. This point, together with the fact that the RoboCup soccer simulator operates in discrete time steps, allows the soccer dribbling task to be mapped onto a discrete-time, episodic reinforcement-learning framework.

Roughly speaking, reinforcement learning is concerned with how an agent must take actions in an environment so as to maximize the expected long-term reward~\cite{Sutton:Reinforcement_learning_book}. Like in a trial-and-error search, the learner must discover which action is the most rewarding one in a given state of the world. Thus, solving a reinforcement learning problem means finding a function (\emph{policy}) that maps \emph{states} to \emph{actions} so that it maximizes a \emph{reward} over the long run. As a way of incorporating domain knowledge, the actions available to the dribbler are the following high-level \emph{macro-actions}, which are built on the simulator's \emph{primitive actions}\footnote{Henceforth, we use the terms \emph{action} and \emph{macro-action} interchangeably, while always distinguishing \emph{primitive actions}.}:

\begin{itemize}

    \item \emph{HoldBall}(): The dribbler holds the ball close to its body, keeping it in a position that is difficult for the adversary to gain possession;

    \item \emph{Dribble}($\Theta, k$): The dribbler turns its body towards the global angle $\Theta$, kicks the ball $k$ meters ahead of it, and moves to intercept the ball.

\end{itemize}

The global angle $\Theta$ is in the range $[0, 360]$. In detail, the center of the training field has been chosen as the origin of the system, where the zero-angle points towards the middle of the right line that delimits the training field, and it increases in the clockwise direction. Those macro-actions are based on high-level skills used by the UvA Trilearn 2003 team~\cite{uva}. The first one maps directly onto the primitive action \emph{kick}. Consequently, it usually takes a single time step to be performed. The second one, however, requires an extended sequence of the primitive actions \emph{turn}, \emph{kick}, and \emph{dash}. To handle this situation, we treat the soccer dribbling task as a \emph{semi-Markov decision process} (SMDP) \cite{Puterman:MDP}.

Formally, an SMDP is a 5-tuple $<S, A, P, r, F>$, where $S$ is a countable set of states, $A$ is a countable set of actions, $P(s^{\prime}|s, a)$, for $s^{\prime}, s \in S$, and $a \in A$, is a probability distribution providing the transition model between states, $r(s, a) \in \Re$ is a reward associated with the transition $(s, a)$, and $F( \tau | s, a)$ is a probability distribution indicating the sojourn time in a given state $s \in S$, \textit{i.e.}, the time before transition provided that action $a$ was taken in state $s$.

Let $a_i \in A$ be the $i$th macro-action selected by the dribbler. Thus, several simulator's time steps may elapse between $a_i$ and $a_{i+1}$. Let $s_{i+1} \in S$ and $r_{i+1} \in \Re$ be, respectively, the state and the reward following the macro-action $a_{i}$. From the dribbler's point of view, an episode consists of a sequence of SMDP steps, \textit{i.e.}, a sequence of states, macro-actions, and rewards: $s_0, a_0, r_1, s_1, \dots, s_i, a_i, r_{i+1}, s_{i+1}, \dots, a_{n-1}, r_n, s_n$, where $a_i$ is chosen based exclusively on the state $s_i$, and $s_n$ is a terminal state in which either the adversary or the dribbler is declared the winner of the episode by the coach. In the former\break case, the dribbler receives the reward $r_n = -1$, while in the latter case its reward is $r_n = 1$. The intermediate rewards are\break always equal to zero, \textit{i.e.}, $r_1 = r_2 = \dots = r_{n-1} = 0$. Thus, our objective is to find a policy that maximizes the dribbler's reward, \textit{i.e.}, the number of episodes in which it is the winner.

\subsection{Dribbler}

The dribbler must take a decision at each SMDP step by selecting an available macro-action. Besides the macro-action \emph{HoldBall}, the set of  actions available to the dribbler contains four instances of the macro-action \emph{Dribble}: \emph{Dribble}($30^\circ, 5$), \emph{Dribble}($330^\circ, 5$), \emph{Dribble}($0^\circ, 5$), and \emph{Dribble}($0^\circ, 10$). Thus, besides hiding the ball from the adversary, the dribbler can kick the ball forward (strongly and weakly), diagonally upward, and diagonally downward. If at some time step the dribbler has not possession of the ball and the current state is not a terminal state, then it usually means that the dribbler chose an instance of the macro-action \emph{Dribble} before and it is currently moving to intercept the ball.

We turn now to the state representation used by the dribbler. It consists of a set of state variables which are based on information related to the ball, the adversary, and the dribbler itself. Let $ang(x)$ be the global angle of the object $x$, and $ang(x, y)$ and $dist(x, y)$ be, respectively, the relative angle and the distance between the objects $x$ and $y$. Further, let $w$ and $h$ be, respectively, the width and the height of the training field. Finally, let $posY(x)$ be a function indicating whether the object $x$ is close to (less than 1 meter away from) the top line or the bottom line that delimits the training field. In the former case, $posY(x) = 1$, whereas in the latter case $posY(x) = -1$, and otherwise $posY(x) = 0$. Table 1 shows the state variables together with their ranges.

\begin{table}[t]
\begin{center}
\renewcommand{\arraystretch}{0.9}
\caption{Description of the state representation.}
\begin{tabular}{|c|c|}
\hline
State Variable &  Range\\
\hline
$posY(\mbox{dribbler})$ & $\{-1, 0 ,1\}$\\
$ang(\mbox{dribbler})$ & $[0, 360]$\\
$ang(\mbox{dribbler}, \mbox{adversary})$ & $[0, 360]$  \\
$ang(\mbox{ball}, \mbox{adversary})$ & $[0, 360]$  \\
$dist(\mbox{ball}, \mbox{adversary})$ & $[0, \sqrt{w^2 + h^2}]$\\
\hline
\end{tabular}
\end{center}
\end{table}

The first three variables help the dribbler to locate itself and the adversary inside the training field. Together, the last two variables can be seen as a point describing the position of the adversary in a polar coordinate system, where the ball is the pole. Thus, these variables are used by the dribbler to locate the adversary with respect to the ball. It is interesting to note that a more informative state representation can be used by adding more state variables, \textit{e.g.}, the current speed of the ball and the dribbler's stamina. However, large domains can be impractical due to the ``curse of dimensionality", \textit{i.e.}, the general tendency of the state space to grow exponentially in the number of state variables~\cite{cursedimensionality}. Consequently, we focus on a state representation that is as concise as possible.

\subsection{Adversary}

The adversary uses a fixed, pre-specified policy. Thus, we can see it as part of the environment in which the dribbler is interacting with. When the adversary has possession of the\break ball, it tries to maintain possession for another time step by\break invoking the macro-action \emph{HoldBall}. If it maintains possession for two consecutive time steps, then it is the winner of the episode. When the adversary does not have the ball, it uses an iterative scheme to compute a near-optimal interception point based on the ball's position and velocity. Thereafter, the adversary moves to that point as fast as possible. This procedure is the same used by the dribbler when it is moving to intercept the ball after invoking the macro-action \emph{Dribble}. More details about this iterative scheme can be found in the description of the UvA Trilearn 2003 team~\cite{uva}.

\section{The Reinforcement Learning Algorithm}

Our solution to the soccer dribbling task combines the reinforcement learning algorithm Sarsa with CMAC for function approximation. In what follows, we briefly introduce both of them before presenting the final learning algorithm.

\subsection{Sarsa}

The Sarsa algorithm works by estimating the action-value function $Q^\pi(s, a)$, for the current policy $\pi$ and for all state-action pairs $(s, a)$~\cite{Sutton:Reinforcement_learning_book}. The $Q$-function assigns to each state-action pair the expected return from it. Given a quintuple of events, $(s_t, a_t, r_{t+1}, s_{t+1}, a_{t+1})$, that make up the transition from the state-action pair $(s_t, a_t)$ to the next one, $(s_{t+1}, a_{t+1})$, the $Q$-value of the first state-action pair is updated according to the following equation:

\begin{equation}
Q(s_t,a_t) \leftarrow Q(s_t,a_t) + \alpha\delta_t,
\end{equation}

\noindent where $\delta_t$ is the traditional \emph{temporal-difference error},

\begin{equation}
\delta_t = r_{t+1} + \lambda Q(s_{t+1},a_{t+1}) - Q(s_t,a_t),
\end{equation}

\noindent $\alpha$ is the learning rate parameter, and $\lambda$ is a discount rate governing the weight placed on future, as opposed to immediate, rewards. Sarsa is an on-policy learning method, meaning that it continually estimates $Q^\pi$, for the current policy $\pi$, and at the same time changes $\pi$ towards greediness with respect to $Q^\pi$. A typical policy derived from the $Q$-function is an \emph{$\epsilon$-greedy} policy. Given the state $s_t$, this policy selects a random action with probability $\epsilon$ and, otherwise, it selects the action with the highest estimated value, \textit{i.e.}, $a = \argmax_a Q(s_t, a)$.

\subsection{CMAC}

In tasks with a small number of state-action pairs, we can represent the action-value function $Q^\pi$ as a table with one entry for each state-action pair. However, this is not the case of the soccer dribbling task. For illustration's sake, suppose that all variables in Table 1 are discrete. If we consider the $5$ actions available to the dribbler and a 20m x 20m training field, we end up with more than $1.9 \times 10^{10}$ state-action pairs. This would not only require an unreasonable amount of memory, but also an enormous amount of data to fill up the table accurately. Thus, we need to \emph{generalize} from previously experienced states to ones that have never been seen. For dealing with this task, we use a technique commonly known as \emph{function approximation}.

By using a function approximation, the action-value function $Q^\pi$ is now represented as a parameterized functional form~\cite{Sutton:Reinforcement_learning_book}. Now, whenever we make a change in one parameter value, we also change the estimated value of many state-action pairs, thus obtaining generalization. In this work, we use the Cerebellar Model Arithmetic Computer (CMAC) for function approximation~\cite{Albus:Tile_Coding_1,Albus:Tile_Coding_2}.

CMAC works by partitioning the state space into multi-dimensional \emph{receptive fields}, each of which is associated with a \emph{weight}. In this work, receptive fields are  hyper-rectangles in the state space. Nearby states share receptive fields. Thus, generalization occurs between them. Multiple partitions of the state space (\emph{layers}) are usually used, which implies that any input vector falls within the range of multiple \emph{excited receptive fields}, one from each layer.

Layers are identical in organization, but each one is offset relative to the others so that each layer cuts the state space in a different way. By overlapping multiple layers, it is possible to
achieve quick generalization while maintaining the ability to learn fine distinctions. Figure 2 shows an example of two grid-like layers overlaid over a two-dimensional space.

\begin{figure}[t]
\centerline{\includegraphics[scale=0.35]{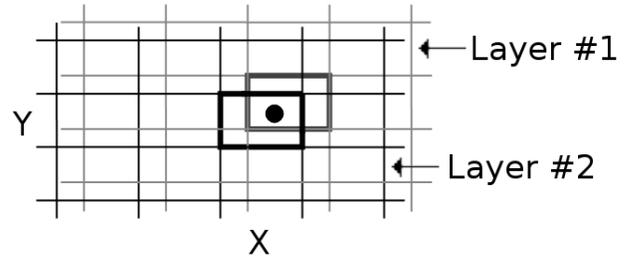}}
\caption{Example of two layers overlaid over a two-dimensional state space. Any input vector (state) activates two receptive fields, one from each layer. For example, the state represented by the black dot activates the highlighted receptive fields.}
\end{figure}

The receptive fields excited by a given state $s$ make up the feature set $\mathbb{F}_s$, with each action $a$ indexing their weights in a different way. In other words, each macro-action is associated with a particular CMAC. Clearly, the number of receptive fields inside each feature set is equal to the number of layers. The CMAC's response to a feature set $\mathbb{F}_s$  is equal to the sum of the weights of the receptive fields in $\mathbb{F}_s$. Formally, let $\theta_a(i)$ be the weight of the receptive field $i$ indexed by the action $a$. Thus, the CMAC's response to $\mathbb{F}_s$ is equal to $\sum_{i \in \mathbb{F}_s} \theta_a(i)$, which represents the $Q$-value $Q(s,a)$.

CMAC is trained by using the traditional \emph{delta rule} (also known as the least mean square). In detail, after selecting an action $a$, the weight of an excited receptive field $i$ indexed by $a$, $\theta_a(i)$, is updated according to the following equation:

\begin{equation}
\theta_{a}(i) \leftarrow \theta_a(i)  +\alpha\delta,
\end{equation}

\noindent where $\delta$ is the temporal-difference error. A major issue when using CMAC is that the total number of receptive fields required to span the entire state space can be very large. Consequently, an unreasonable amount of memory may be needed. A technique commonly used to address this issue is called \emph{pseudo-random hashing}~\cite{Sutton:Reinforcement_learning_book}. It produces receptive fields consisting of noncontiguous, disjoint regions randomly spread throughout the state space, so that only information about receptive fields that have been excited during previous training is actually stored.

\subsection{Linear, Gradient-Descent Sarsa}

Our solution to the soccer dribbling task combines the Sarsa algorithm with CMAC for function approximation. We use an $\epsilon$-greedy policy for action selection. Sutton and Barto~\cite{Sutton:Reinforcement_learning_book} provide a complete description of this algorithm under the name of \emph{linear, gradient-descent Sarsa}. Our implementation follows the solution proposed by Stone~\textit{et al.}~\cite{Stone:Keepaway}. It consists of three routines: \emph{RLstartEpisode}, to be run by the dribbler at the beginning of each episode; \emph{RLstep}, run on each SMDP step; and \emph{RLendEpisode}, to be run when an episode ends. In what follows, we present each routine in detail.

\subsubsection{RLstartEpisode}

Given an initial state $s_0$, this routine starts by iterating over all available actions. In line 2, it finds the receptive fields excited by $s_0$, which compose the feature set $\mathbb{F}_{s_0}$. Next, in line 3, the estimated value of each macro-action $a$ in $s_0$ is calculated as the sum of the weights of the excited receptive fields. In line 5, this routine selects a macro-action by following an $\epsilon$-greedy policy and sends it to the RoboCup soccer simulator. Finally, the chosen action and the initial state $s_0$ are stored, respectively, in the variables $LastAction$ and $LastState$.

\begin{algorithm}[H]
\caption{RLstartEpisode}
\begin{algorithmic}[1] 
\FOR{ each action $a$} 
\STATE $\mathbb{F}_{s_0} \leftarrow $ receptive fields excited by $s_0$
\STATE $Q_a \leftarrow \sum_{i \in \mathbb{F}_{s_0}} \theta_a(i)$
\ENDFOR
\STATE $
LastAction \leftarrow \left\{ \begin{array}{ll}
\argmax_a Q_a & \mbox{w/ prob. } 1 - \epsilon \\
\mbox{random action} & \mbox{w/ prob. } \epsilon \\
\end{array} \right.
$
\STATE $LastState \leftarrow s_0 $
\end{algorithmic}
\end{algorithm}

\subsubsection{RLstep}

This routine is run on each SMDP step, whenever the dribbler has to choose a macro-action. Given the current state $s$, it starts by calculating part of the temporal-difference error (Equation 2), namely the difference between the intermediate reward $r$ and the expected return of the previous SMDP step, $Q_{LastAction}$. In lines 2 to 5, this routine finds the receptive fields excited  by $s$ and uses their weights to compute the estimated value of each action $a$ in $s$. In line 6, the next action to be taken by the dribbler is selected according to an $\epsilon$-greedy policy. In line 7, this routine finishes to compute the temporal-difference error by adding the discount rate $\lambda$ times the expected return of the current SMDP step, $Q_{CurrentAction}$. Next, in lines 8 to 10, this routine adjusts the weights of the receptive fields excited in the previous SMDP step by the learning factor $\alpha$ times the temporal-difference error $\delta$ (see Equation 3). Since the weights have changed, we must recalculate the expected return of the current SMDP step, $Q_{CurrentAction}$ (line 11). Finally, the chosen action and the current state are stored, respectively, in the variables $LastAction$ and $LastState$.

\begin{algorithm}[H]
\caption{RLstep}
\begin{algorithmic}[1] 
\STATE $\delta \leftarrow r - Q_{LastAction}$
\FOR{ each action $a$} 
\STATE $\mathbb{F}_{s}  \leftarrow $ receptive fields excited by $s$
\STATE $Q_a \leftarrow \sum_{i \in \mathbb{F}_{s}} \theta_a(i)$
\ENDFOR
\STATE $
CurrentAction \leftarrow \left\{ \begin{array}{ll}
\argmax_a Q_a & \mbox{w/ prob. } 1 - \epsilon \\
\mbox{random action} & \mbox{w/ prob. } \epsilon \\
\end{array} \right.$
\STATE $\delta \leftarrow \delta + \lambda Q_{CurrentAction}$
\FOR{ each $i \in \mathbb{F}_{LastState}$} 
\STATE $\theta_{LastAction}(i) \leftarrow \theta_{LastAction}(i) + \alpha\delta $
\ENDFOR
\STATE $Q_{CurrentAction} \leftarrow \sum_{i \in \mathbb{F}_{s}} \theta_{CurrentAction}(i)$
\STATE $LastAction \leftarrow CurrentAction$
\STATE $LastState \leftarrow s$
\end{algorithmic}
\end{algorithm}

\subsubsection{RLendEpisode}

This routine is run when an episode ends. Initially, it calculates the appropriate reward based on who won the episode. Next, it calculates the temporal-difference error in the action-value estimates (line 6). There is no need to add the expected return of the current SMDP step ($Q_{CurrentAction}$) since this value is defined to be $0$ for terminal states. Lastly, this routine adjusts the weights of the receptive fields excited in the previous SMDP step.

\begin{algorithm}[H]
\caption{RLendEpisode}
\begin{algorithmic}[1] 
\IF{the dribbler is the winner}
\STATE $r \leftarrow 1$
\ELSE
\STATE $r \leftarrow -1$
\ENDIF
\STATE $\delta \leftarrow r - Q_{LastAction}$
\FOR{ each $i \in \mathbb{F}_{LastState}$} 
\STATE $\theta_{LastAction}(i) \leftarrow \theta_{LastAction}(i) + \alpha\delta $
\ENDFOR
\end{algorithmic}
\end{algorithm}

\section{Empirical Results}

In this section, we report our experimental results with the soccer dribbler task. In all experiments, we used the standard RoboCup soccer simulator (version 14.0.3, protocol 9.3) and a 20m x 20m training region. In that simulator, agents typically have limited and noisy visual sensors. For example, each player can see objects within a $90^\circ$ view cone, and the precision of an object's sensed location degrades with distance. To simplify the learning process, we removed those restrictions. Both the dribbler and the adversary were given $360^\circ$ of noiseless vision to ensure that they would always have complete and accurate knowledge of the environment.

Related to parameters of the reinforcement learning algorithm\footnote{The implementation of the learning algorithm can be found at:\break \url{http://sites.google.com/site/soccerdribbling/}}, we set $\epsilon = 0.01$, $\alpha = 0.125$, and $\lambda = 1$. By no means do we argue that these values are optimal. They were set based on results of brief, informal experiments.

The weights of first-time excited receptive fields were set to $0$. The bounds of the receptive fields were set according to the generalization that we desired: angles were given widths of about 20 degrees, and distances were given widths of approximately 3 meters. We used 32 layers. Each dimension of every layer was offset from the others by $1/32$ of the desired width in that dimension. We used the CMAC implementation proposed by Miller and Glanz~\cite{cmac_code}, which uses pseudo-random hashing. To retain previously trained information in the presence of subsequent novel data, we did not allow hash collisions.

To create episodes as realistic as possible, agents were not allowed to recover their staminas by themselves. This task was done by the coach after five consecutive episodes. This enabled agents to start episodes with different stamina values. We ran this experiment 5 independent times, each one lasting 50,000 episodes, and taking, on average, approximately 74 hours. Figure 3 shows the histogram of the average number of episodes won by the dribbler during the training process. Bins of  500 episodes were used.

Throughout the training process, the dribbler won, on average, $23,607$ episodes ($\approx 47\%$). From Figure 3, we can see that it greatly improves its average performance as the number of episodes increases. At the end of the training process, it is winning slightly less than $53\%$ of the time.

Qualitatively, the dribbler seems to learn two major rules. In the first one, when the adversary is at a considerable distance, the dribbler keeps kicking the ball to the opposite side in which the adversary is located until the angle between them is in the range $[90, 270]$, \textit{i.e.}, when the adversary is behind the dribbler. After that, the dribbler starts to kick the ball\break forward. An illustration of this rule can be seen in Figure 4.

The second rule seems to occur when the adversary is relatively close to and in front of the dribbler. Since there is no way for the dribbler to move forward or diagonally without putting the possession at risk, it then holds the ball until the angle between it and the adversary is in the range $[90, 270]$. Thereafter, it starts to advance by kicking the ball forward. An illustration of this rule can be seen in Figure 5.

After the training process, we randomly generated 10,000 initial configurations to test our solution. This time, the dribbler always selected the macro-action with the highest estimated value, \textit{i.e.}, we set $\epsilon = 0$. Further, the weights of the receptive fields were not updated, \textit{i.e.}, we set $\alpha = 0$.  We used the receptive fields' weights resulting from the simulation where the dribbler obtained the highest success rate. The result of this experiment was even better. The dribbler won 5,795 episodes, thus obtaining a success rate of approximately $58\%$.

\begin{figure}[t]
\centerline{\includegraphics[scale=0.26]{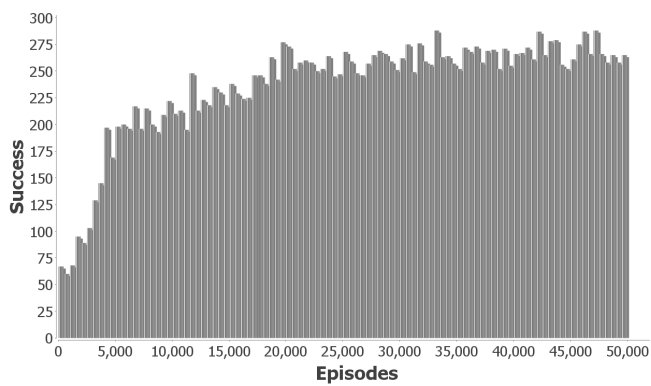}}
\caption{Histogram of the average number of episodes won by the dribbler (success) during the training process (50,000 episodes). Bins of  500 episodes were used, and 5 independent simulations were performed.}
\end{figure}

\begin{figure}[H]
\centerline{\includegraphics[scale=0.46]{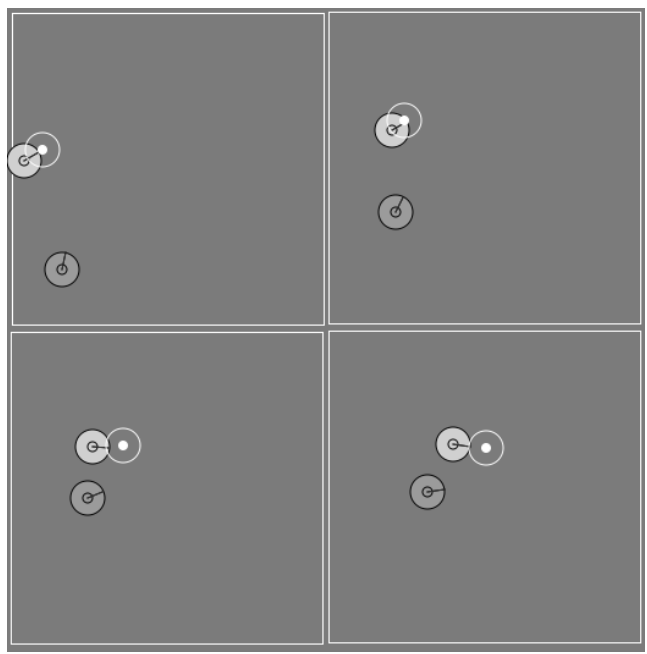}}
\caption{Example of the first major rule learned by the dribbler. (Top Left) The adversary is at a considerable distance from the dribbler. (Top Right) The dribbler starts to kick the ball to the opposite side in which the adversary is located. (Bottom Left) The angle between the adversary and the dribbler is in the range $[90, 270]$. Consequently, the dribbler starts to kick the ball forward. (Bottom Right) The dribbler keeps kicking the ball forward.}
\end{figure}

\begin{figure}
\centerline{\includegraphics[scale=0.46]{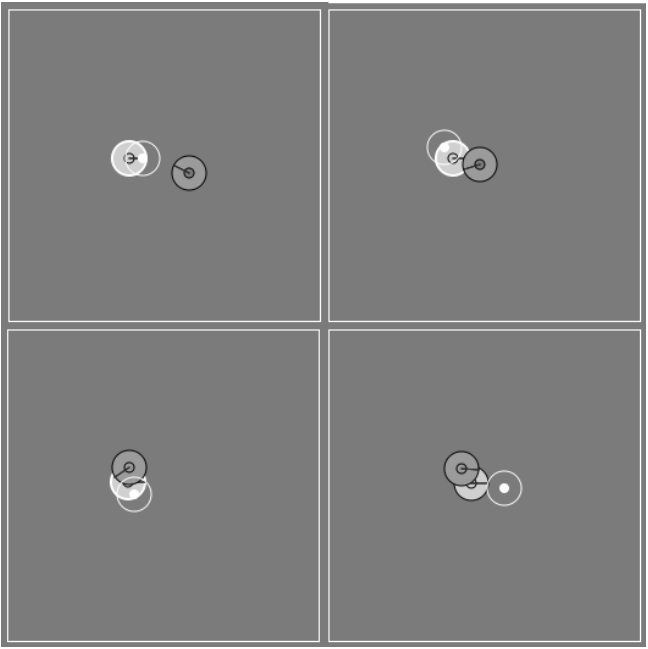}}
\caption{Example of the second major rule learned by the dribbler. (Top Left) The adversary is close to and in front of the dribbler. (Top Right) The dribbler holds the ball so as not to lose possession. (Bottom Left) The dribbler keeps holding the ball. (Bottom Right) The angle between the adversary and the dribbler is the range $[90, 270]$. Consequently, the dribbler starts to advance by kicking the ball forward.}
\end{figure}

\subsection{One-Dimensional CMACs}

For comparison's sake, we repeated the above experiment using the original solution proposed by Stone \textit{et al.}~\cite{Stone:Keepaway}. It consists of the same learning algorithm presented in Section 3, but using one-dimensional CMACs. In detail, each layer is an interval along a state variable. In this way, the feature set $\mathbb{F}_s$ is now composed by $32 \times 5 = 160$ excited receptive fields, \textit{i.e.}, $32$ excited receptive fields for each state variable.

One of the main advantages of using one-dimensional CMACs is that it is possible to circumvent the curse of dimensionality. In detail, the state space does not grow exponentially in the number of state variables because dependence between variables is not taken into account.

Figure 6 shows the histogram of the average number of episodes won by the dribbler during the training process. Each simulation took, on average, approximately 43 hours. Throughout the training process, the dribbler won, on average, $16,278$ episodes ($\approx 33\%$). From Figure 6, we can see that the learning algorithm converges much faster when using one-dimensional CMACs. However, its average performance is considerably worse. At the end of the training process, the dribbler is winning, on average, less than $30\%$ of the time.

After the training process, we tested this solution using the same 10,000 initial configurations previously generated. Again,  we set $\epsilon = \alpha = 0$, and used the receptive fields' weights resulting from the simulation where the dribbler obtained the highest success rate. The result of this experiment was slightly better. The dribbler won 3,701 episodes, thus obtaining a success rate of approximately $37\%$.

\begin{figure}
\centerline{\includegraphics[scale=0.26]{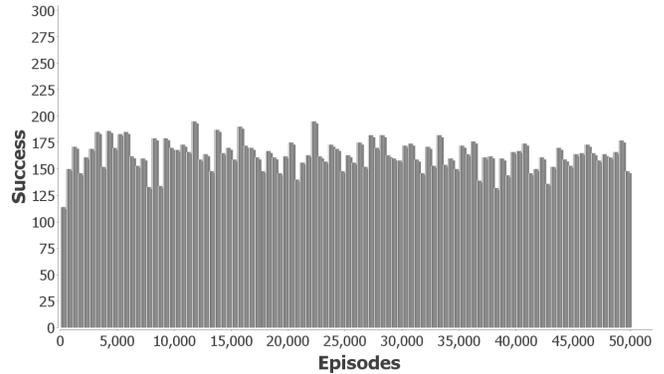}}
\caption{Histogram of the average number of episodes won by the dribbler (success) during the training process (50,000 episodes) when using one-dimensional CMACs. Bins of  500 episodes were used, and 5 independent simulations were performed.}
\end{figure}

Qualitatively, the dribbler seems to learn a rule similar to the one shown in Figure 4. The major difference is that it always kicks the ball to the opposite side in which the adversary is located, it does not matter its distance from the adversary's location. Consequently, it is highly unlikely that the dribbler succeeds when the adversary is close to it.

We conjecture that one of the main reasons for such a poor performance of the reinforcement learning algorithm when using one-dimensional CMACs is that it does not take into account dependence between variables, \textit{i.e.}, they are treated individually. Hence, such approach may throw away valuable information. For example, the variables $ang(\mbox{ball}, \mbox{adversary})$ and $dist(\mbox{ball}, \mbox{adversary})$ together describe the position of adversary with respect to the ball. However, they do not make as much sense when considered individually.

\section{Related Work}

Reinforcement learning has long been applied to the robot soccer domain. For example, Andou~\cite{Andou:Reinforcement_Learning_Robocup} uses ``observational reinforcement learning" to refine a function that is used by\break the soccer agents for deciding their positions on the field. Riedmiller \textit{et al.} \cite{Riedmiller:Reinforcement_Learning_Robocup} use reinforcement learning to learn low-level soccer skills, such as kicking and ball-interception. Nakashima \textit{et al.}~\cite{fuzzyQlearning} propose a reinforcement learning meth-od called ``fuzzy Q-learning", where an agent determines its\break action based on the inference result of a fuzzy rule-based system.  The authors apply the proposed method to the sce-nario where a soccer agent learns to intercept a passed ball.

Arguably, the most successful application is due to Stone \textit{et al.}~\cite{Stone:Keepaway}. They propose the ``keepaway task", which consists of two teams, the keepers and the takers, where the former tries to keep control of the ball for as long as possible, while the latter tries to gain possession. Our solution to the soccer dribbling task follows closely the solution proposed by those\break authors to learn the keepers' behavior. Iscen and Erogul \cite{Iscen:takes} use similar solution to learn a policy for the takers.

Gabel \textit{et al.}~\cite{NeuroHassle} propose a task which is the opposite of the soccer dribbling task, where a defensive player must interfere and disturb the opponent that has possession of the ball. Their solution to that task uses a reinforcement learning algorithm with a multilayer neural network for function approximation.

Kalyanakrishnan \textit{et al.}~\cite{Stone:half_field} present the ``half-field offense task", a scenario in which an offense team attempts to outplay a defense team in order to shoot
goals. Those authors pose that task as a reinforcement learning problem, and propose a new learning algorithm for dealing with it.

More closely related to our work are reinforcement learn-ing-based solutions to the task of conducting the ball (\textit{e.g.}, \cite{riedmiller2008learning}), which can be seen as a simplification of the dribbling task since it usually does not include adversaries.

\section{Conclusion}

We proposed a reinforcement learning solution to the soccer dribbling task, a scenario in which an agent has to go from the beginning to the end of a region keeping possession of the ball, while an adversary attempts to gain possession. Our solution combined the Sarsa algorithm with CMAC for function approximation. Empirical results showed that, after the training period, the dribbler was able to accomplish its task against a strong adversary around $58\%$ of the time.

Although we restricted ourselves to the soccer domain, dribbling, as defined in this paper, is also common in other sports, \textit{e.g.}, hockey, basketball, and football. Thus, the proposed solution can be of value to dribbling tasks of other sports games. Furthermore, we believe that the soccer dribbling task is an excellent benchmark for comparing different machine learning techniques because it involves a complex problem, and it has a well-defined objective.

There are several exciting directions for extending this work. From a practical perspective, we intend to analyze the scalability of our solution, \textit{i.e.}, to study how it performs with training fields of distinct sizes and against different adversaries. Further, we are considering schemes to extend our solution to the original partially observable environment, where the available  information is incomplete and noisy.

As stated before, a more informative state representation could be obtained by using more state variables. The major problem of adding extra variables to our solution is that CMAC's complexity increases exponentially with its dimensionality. Due to this fact, we are considering other solutions which use function approximations whose complexity is unaffected by dimensionality \textit{per se}, \textit{e.g.}, the Kanerva coding (for example, see Kostiadis and Hu's work~\cite{kostiadis2001kabage}).

Finally, we note that when modeling the soccer dribbling task as a reinforcement learning problem, we do not directly use intermediate rewards (they are all set to zero). However, they may make the learning process more efficient (for example, see \cite{Ng:policy_invariance}). Thus, we intend to investigate the influence of intermediate rewards on the final solution in future work.

\section*{Acknowledgments}
We would like to thank W. Thomas Miller, Filson H. Glanz, and others from the Department of Electrical and Computer Engineering at the University of New Hampshire for making their CMAC code available.

\bibliographystyle{IEEEtran}
\bibliography{SamplePaper}

\end{document}